\documentclass[11pt,a4paper]{article}
\usepackage[hyperref]{acl2020}
\usepackage{times}
\usepackage{latexsym}

\usepackage{microtype}

\usepackage[ruled,vlined]{algorithm2e}
\usepackage{amsmath}
\usepackage{multirow}
\usepackage{makecell,color}
\usepackage{graphicx}

\makeatletter
\newcommand{\thickhline}{%
    \noalign {\ifnum 0=`}\fi \hrule height 0.8pt
    \futurelet \reserved@a \@xhline
}
\makeatother

\aclfinalcopy 

\title{Table-to-Text Natural Language Generation with Unseen Schemas}

\author{Tianyu Liu$^1$, Wei Wei$^2$, William Yang Wang$^3$  \\
  Peking University$^1$ \\
  Google AI$^2$ \\
  University of California, Santa Barbara$^3$ \\
  \texttt{ty-liu@pku.edu.cn}\\
  \texttt{wewei@google.com} \\ 
  \texttt{william@cs.ucsb.edu} 
  }

\date{}

\begin{document}
\maketitle

\begin{abstract}
    Traditional table-to-text natural language generation (NLG) tasks focus on generating text from schemas that are already seen in the training set. This limitation curbs their generalizabilities towards real-world scenarios, where the schemas of input tables are potentially infinite. In this paper, we propose the new task of table-to-text NLG with unseen schemas, which specifically aims to test the generalization of NLG for input tables with attribute types that never appear during training. To do this, we construct a new benchmark dataset for this task. To deal with the problem of unseen attribute types, we propose a new model that first aligns unseen table schemas to seen ones, and then generates text with updated table representations. Experimental evaluation on the new benchmark demonstrates that our model outperforms baseline methods by a large margin. In addition, comparison with standard data-to-text settings shows the challenges and uniqueness of our proposed task.
\end{abstract}

\section{Introduction}

\begin{figure}
        \centering
        \includegraphics[scale=0.65]{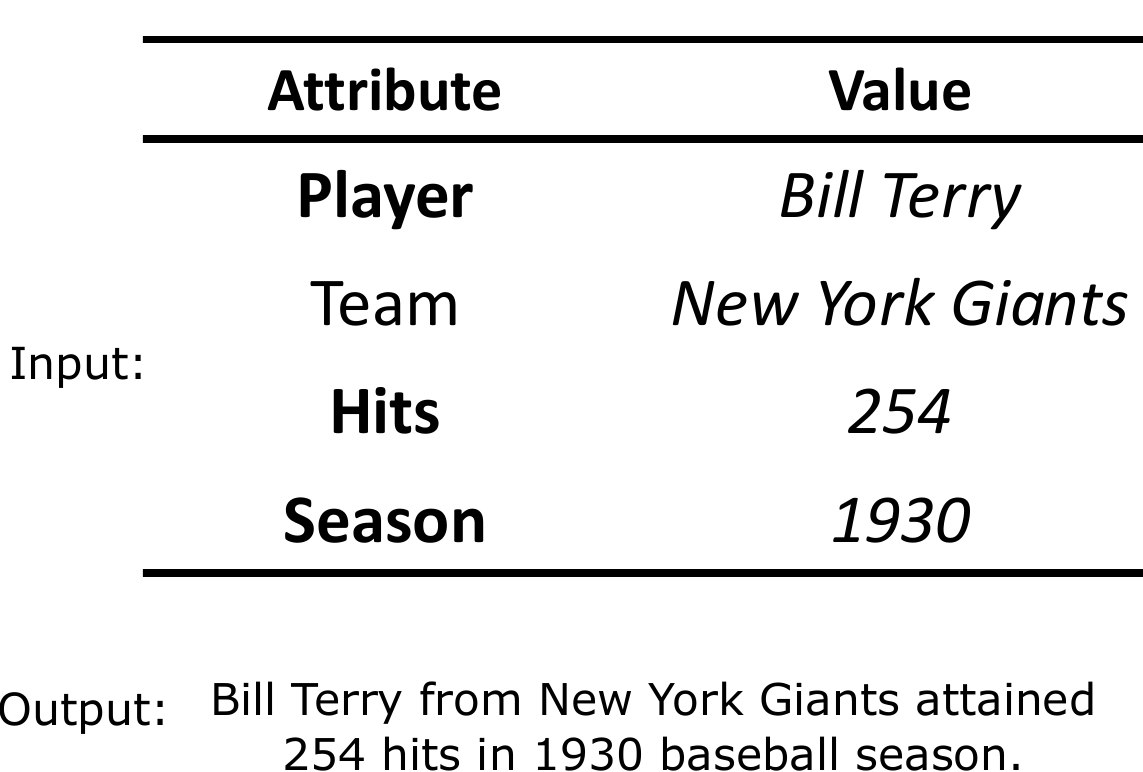}
        \caption{An example of table-to-text generation from Wikipedia. Given a table as the input, we aim to generate a description of the contents of the table. In this work, we tackle a critical challenge in table-to-text generation unseen attribute types (highlighted in bold) appear in the testing scenarios.}
        \label{fig:example}
    \end{figure}

    Over the past few years, table-to-text natural language generation has received increasingly more attention. Typically, table-to-text generation model takes in a table as input and aims to generate a description of its content in natural language as in the example given in figure \ref{fig:example}. A table consists of several \textit{attribute}-\textit{value} pairs. In this work, we refer to the combination of various attributes that appear in one table as \textit{schema}. The task of table-to-text generation requires the ability to first understand the information conveyed by the table and then generate fluent natural language to describe the information. Great potential lies in utilizing table-to-text techniques in real-world applications such as question answering, automatic news writing, and task-oriented dialog system.
    
    A wide range of table-to-text tasks and datasets have been proposed over the past few years such as \textsc{weathergov} \cite{belz_2008}, \textsc{wikibio} \cite{lebret-etal-2016-neural}, and the E2E challenge \cite{novikova-etal-2017-e2e}. However, these existing tasks are designed for specific domains with limited attribute types and simple schema patterns. For instance, the E2E dataset \cite{novikova-etal-2017-e2e} constructed for restaurant domain has only 8 predefined attribute types and the \textsc{weathergov} \cite{belz_2008} designed for automatic weather report generation has only 10. A notable exception is the \textsc{wikibio} dataset \cite{lebret-etal-2016-neural} that has approximately 7K attribute types. However, most of its instances follow the template of biography, which puts more emphasis on monotonous information such as names, birth dates, and occupations. Although currently prevalent data-driven end-to-end models \cite{AAAI1816138,AAAI1816599,gardent-etal-2017-webnlg} yield promising results on these tasks, they implicitly bias towards fixed schema-utterance pairs. Such bias limits their generalizabilities towards real-world scenarios where various unseen attribute types may appear in the input tables. 
    
    Therefore, we introduce the new task of \textit{table-to-text generation for unseen schemas}, where we focus on generating descriptions for input table with schemas that never appear during training. Since previous close-domain table-to-text tasks has little room for generalization, we choose to conduct our experiments on the open-domain dataset \textsc{wikitabletext} \cite{AAAI1816138} collected from the entire Wikipedia without being restricted to any specific domains. The abundance and diversity of its attribute types and schemas make it possible for us to control the proportion of unseen attributes during the testing phase through subsampling the training set and construct a proper benchmark for our task.

    In parallel, conventional methods \cite{Konstas:2013:GMC:2591248.2591256,moryossef-etal-2019-step,ma-etal-2019-key} deal with table-to-text task in a two-stage (namely content planning and surface realization) manner. By such means, table understanding and text generation are separated, and the result of text generation is explicitly conditioned on the result of content planning. This quality paves the way to tackling with unseen input table schemas by controlling over their representations while avoiding undermining the reliability of the text generation part.
    
    In light of the points raised above, we propose a novel table-to-text model, AlignNet, which explicitly learns an alignment between unseen schemas with seen ones. Our method is by nature a two-stage model while maintaining the property to be trained end-to-end. When the model receives an unseen table schema as input, it first infers possible alignments with seen schemas in the train set. Then, representations of unseen attribute types are replaced with best aligned seen ones. In a nutshell, our work has the following contributions:
    \begin{itemize}
        \item We propose the new task of table-to-text natural language generation for unseen schemas. Compared with traditional table-to-text tasks, the new task is closer to real-world scenarios.
        
        \item In order to deal with the new problem, we propose a novel end-to-end neural model that explicitly learns to align unseen schemas to seen ones.
        
        \item We construct a benchmark dataset for this new task and demonstrate the effectiveness and capability of our method to deal with unseen table schemas.
    \end{itemize}

\section{Related Work}
    
    \subsection{Data-to-Text Generation}
        Data-to-text generation is a vibrant subdomain of natural language generation, alongside a wide range of fields such as machine translation \cite{kalchbrenner-blunsom-2013-recurrent-continuous,bahdanau2014neural}, document summarization \cite{rush-etal-2015-neural,gu-etal-2016-incorporating,see-etal-2017-get}, and dialog system \cite{DBLP:journals/corr/VinyalsL15,shang-etal-2015-neural,li-etal-2016-deep}. The speciality of table-to-text generation is, by definition, non-linguistic input \cite{reiter_dale_1997,Konstas:2013:GMC:2591248.2591256}. Traditionally, data-to-text generation systems are implemented in a two-stage manner. The first core step determines what to say (content planning) and the second step determines how to say (surface realization). Earlier surface realization models focus on generating natural language from rules and hand-crafted templates \cite{Reiter:2005:CWC:1113166.1644548,Dale:2003:CUN:783106.783111}, meaning representation language (MRL) \cite{wong-mooney-2007-generation}, probabilistic context-free grammar (PCFG) \cite{Cahill:2006:RPG:1220175.1220305}, etc. Later, end-to-end unified models \cite{angeli-etal-2010-simple,Konstas:2013:GMC:2591248.2591256} that combine the two steps by joint optimization became prevalent.
        
        The trend of merging content planning and surface realization became even stronger since the introduction of sequence-to-sequence framework \cite{Sutskever:2014:SSL:2969033.2969173}. A number of improvements upon sequence-to-sequence framework such as copy mechanism \cite{AAAI1816138}, attention mechanism \cite{AAAI1816599}, symbolic reasoning \cite{nie-etal-2018-operation} have been explored. In the meantime, attempts \cite{DBLP:journals/corr/abs-1809-00582} that aim to decompose sequence-to-sequence framework into content planning and surface realization without sacrificing the end-to-end trainable property have yielded promising results. Nevertheless, most previous studies are conducted in closed-world settings that focus on limited input attribute types and schemas and pay little attention to generalizability.

    \subsection{Domain Adaptation and Zero-Shot Learning}
        Domain adaptation typically involves adapting models trained on rich-resource domains to low-resource domains. Recent years have seen growing efforts on domain adaptation for NLG tasks, such as machine translation \cite{hu-etal-2019-domain}, dialog systems \cite{qian-yu-2019-domain}. In terms of data-to-text generation, \citeauthor{angeli-etal-2010-simple} \shortcite{angeli-etal-2010-simple} first propose a unified domain-independent framework that does not require domain-specific feature engineering. \citeauthor{wen-etal-2016-multi} \shortcite{wen-etal-2016-multi} manually create ontologies for different domains and leverage data augmentation technique to adapt a data-to-text generation system to multiple domains. 
        
        Meanwhile, zero-shot learning can be regarded as a special case of transfer learning, where no label information of the target domain can be obtained during learning. In prior work, zero-shot learning has been studied for question generation \cite{elsahar-etal-2018-zero}, image captioning \cite{DBLP:journals/corr/abs-1811-02765}, dialog generation \cite{zhao-eskenazi-2018-zero}, etc.
        
        Our proposed task bears some similarities to domain adaption and zero-shot learning in the way that the evaluation is conducted on test sets which contain unseen attribute types and the distribution of data during the testing phase is different from that during the training phase.
        
        On the other hand, the proposed task is different from standard domain adaptation since we do not explicitly define domains such as restaurant, sports, biography in our task. Our main focus is to evaluate the generalizability towards any given schema rather than another specific domain. Thus, it is not necessary to obtain the domain-specific knowledge from external resources that is required in most zero-shot learning and domain adaptation methods \cite{wen-etal-2016-multi,zhao-eskenazi-2018-zero}.
        

        \begin{figure*}[!ht]
            \centering
            \includegraphics[scale=0.7]{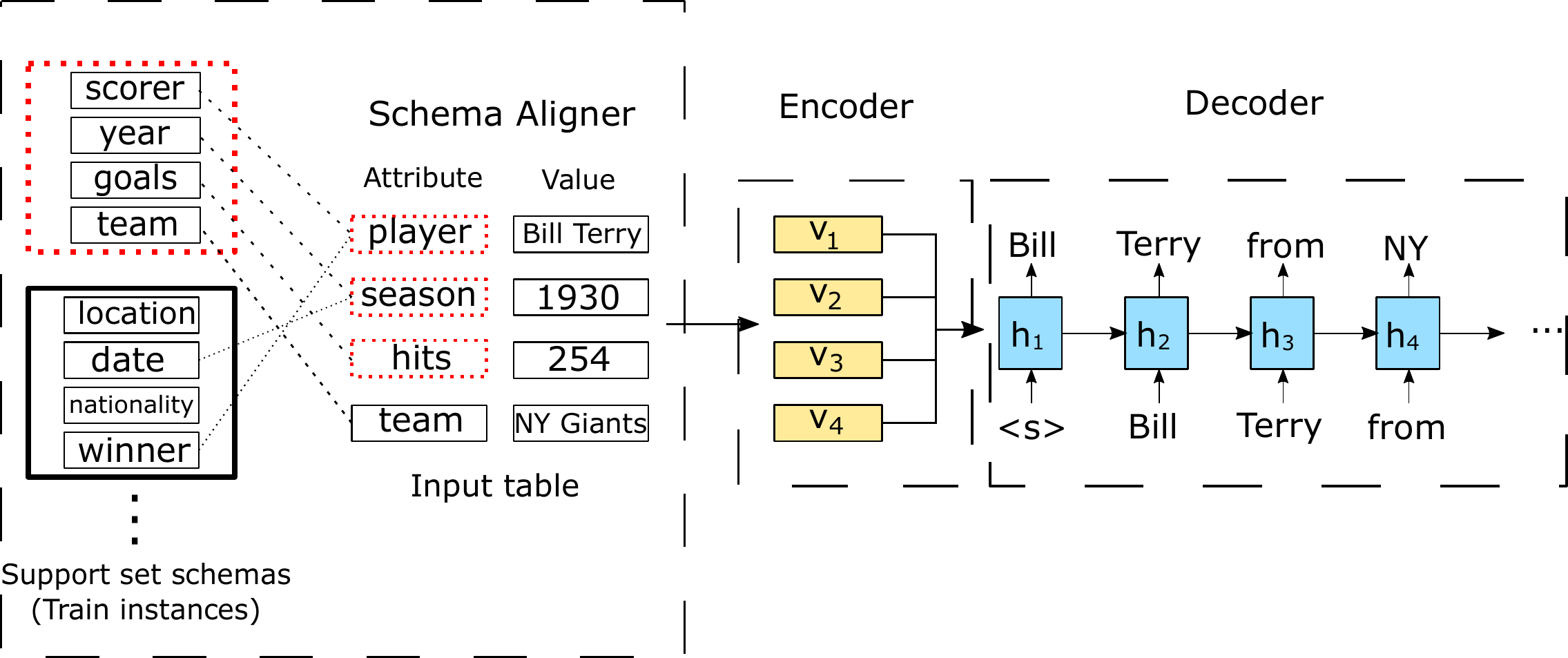}
            \caption{The overall architecture of our model. Attention and copy mechanism are omitted for simplicity. Unseen attribute types (player, season, hits) of the input table are in dotted red boxes. The schema aligner first computes alignment scores for each support set schema and select the schema with the highest alignment score (the one in dotted red box) to replace the paired attribute representations. The encoder then takes in the updated representations as input. During training, we sample the support set from the training set excluding the current training instance. During testing, the support set is sampled from the whole training set.}
            \label{fig:alignnet}
        \end{figure*}

\section{Task Definition}
    In this section, we first describe the formalization of general table-to-text tasks and then introduce the new task of table-to-text generation for unseen schemas.

    \subsection{Formalization of Table-to-Text}
        Provided with a table $T$ as the input, the table-to-text generation task aims to generate a natural language sequence $\mathbf{y}=\{y_1,y_2,\cdots,y_m\}$ that describes the content of $T$ as output. In this work, a table $T$ is defined as a list of $n$ attribute-value pairs $\{a_i:c_i \}^n_{i=1}$.

    \subsection{Table-to-Text Generation for Unseen Schemas}
        Traditional table-to-text generation tasks and datasets put little emphasis on evaluating the performance on unseen schemas. In real-world applications, however, the types of attributes are not limited to those that appear during training. Thus, these traditional table-to-text tasks has limited generalizability towards real-world scenarios.
        
        In view of the weakness mentioned above, we propose the new task of \textit{table-to-text generation from unseen schemas}. Unlike traditional settings, the new task imitates open-world scenarios. It explicitly aims to generate texts from schemas with a large proportion of attribute types that never appear in the training set.

\section{Our Framework}
        
    \subsection{Table-to-Sequence Framework}
        Our model is based on the state-of-the-art table-to-sequence framework proposed in \cite{AAAI1816138}. Their method is by nature an encoder-decoder model which first encodes an input table into a vector representation and then decodes it into a natural language sequence. In the course of decoding, attention mechanism and copy mechanism are leveraged in order to generate accurate . 
    
        \subsubsection{Table Encoder Module}
        First, the model embeds attribute types and cell contents into vector representations. Specifically, each attribute $a_i$ is represented as a vector $\mathbf{e}_i^a$ and its corresponding content cell $c_i$ is represented as a vector $\mathbf{e}_i^c$. Afterwards, we compute the final representation of an \textit{attribute}-\textit{value} pair using $\mathbf{v}_i = \mathrm{FFN}_e([e_i^a;e_i^c])$, where $[\,;\,]$ is the concatenation operator and $\mathrm{FFN}_e$ is a single-layer feedforward neural network.
        
        Then, a vector $\mathbf{h}_0$ representing the whole table is computed by a sequence-to-vector encoder $\mathbf{h}_0 = f_{enc}(\langle \mathbf{v}_1,\mathbf{v}_2,\cdots,\mathbf{v}_n \rangle)$ and utilized to initialize the decoder state. 
        
        \subsubsection{Decoder Module}
        The model decoder utilizes copy mechanism \cite{gu-etal-2016-incorporating} that is capable of copying tokens from both table attributes and cell contents. At decoding step $t$, an LSTM-based decoder takes in the predicted token representation output $\mathbf{y}_{t-1}$ of step $t-1$, the hidden state $\mathbf{h}_{t-1}$ of step $t-1$, and an attentive vector $\mathbf{m}_t$ as inputs. The recurrence of decoding procedure is given by
    
        \begin{equation}
            h_t = f_{dec}(\mathbf{y}_{t-1},\mathbf{h}_{t-1},\mathbf{m}_t)
        \end{equation}
        \begin{equation}
            \mathbf{m}_t = \sum_{i=1}^{n} \alpha_{ti}\mathbf{v}_i ; \;\;  
            \alpha_{ti} = \frac{exp(\eta(\mathbf{m}_{t-1},\mathbf{v}_i))}{\sum_{j=1}^{n}exp(\eta(\mathbf{m}_{t-1},\mathbf{v}_j))}
        \end{equation}
        
        where $\mathbf{v}_i$ is the representation of the $i$-th table cell and $\eta(\cdot,\cdot)$ is a bilinear attention function. Decoder output $\mathbf{h}_t$ of step $t$ is then fed into a word prediction layer to determine a generation score for each word in the vocabulary.
        
        With regard to copy mechanism, a copy score $g_i^a$ for attribute and a copy score $g_i^c$ for cell content is computed for each \textit{attribute}-\textit{value} pair of the input table $T$ by
        
        \begin{equation}
            \begin{aligned}
                g_i^a &= \mathrm{FFN}_{g^a}([\mathbf{y}_{t-1};\mathbf{h}_t;\mathbf{m}_t;\mathbf{v}_i;\mathbf{e}_i^a]); \\
                g_i^c &= \mathrm{FFN}_{g^c}([\mathbf{y}_{t-1};\mathbf{h}_t;\mathbf{m}_t;\mathbf{v}_i;\mathbf{e}_i^c]) .
            \end{aligned}
        \end{equation}
        
        The generation scores and copy scores are then concatenated and fed into a $\mathrm{softmax}$ function to calculate the final probability distribution over the vocabulary set extended with input table contents.

        \subsection{Table Alignment and Attribute Representation Replacement}
            
            To address the challenge presented by unseen schemas during the test phase, we begin with the intuition that different tables may share some common fields of contents even though they have different schemas. For instance, tables that describe sport game events often use ``season'' while tables that describe general historical events usually use ``year'' to represent the attribute of time. Another example is that tables may use various types of words such as ``player'', ``winner'', ``actor'' in terms of the attribute of \textit{person}. Therefore, it is reasonable to seek for possible paraphrases within the training set when the table-to-text model encounters unseen attribute types during testing.

            \begin{algorithm}[!hb]
            \SetAlgoLined
                \LinesNumbered
                \KwIn{Table $T^S$ from support set with attribute representations $\{\mathbf{v}_{j}^{S}\}_{j=1}^{m}$ ; input table $T$ with attribute representations $\{\mathbf{v}_{i}\}_{i=1}^{n}$}
                \KwOut{Alignment score $r$ ; Alignment $\{(i,j_i)\}_{i=1}^{n}$.}
                
                 Calculate similarity matrix $\mathbf{S}=\{s_{ij}=\frac{\mathbf{v}_{i}\mathbf{v}_{j}^{S}}{\lvert \mathbf{v}_{i} \rvert \lvert \mathbf{v}_{j}^{S} \rvert} \}_{n \times m}$\; 
                 
                 Use the Hungarian Algorithm\footnotemark{} to find an optimal alignment $\{(i,j_i)\}_{i=1}^{n}$ that maximizes the sum of similarity scores\;
                 
                 
                 
                 
                 
                 
                 
                 Discard alignment pairs $\{(i,j_i)\}$ where $s_{i j_i} < 0$; Compute alignment score using $r = \sum_{i=1}^n s_{ij_i}$.
                 
                 \caption{Schema Alignment}\label{alg:align}
            \end{algorithm} 
            
            \footnotetext{The overall complexity of the Hungarian Algorithm is $O(n^3)$ where $n$ is the number of attribute types in a table. Since $n$ is relatively small ($<10$) in our work, the computational cost of algorithm \ref{alg:align} is affordable.}

            In this section, we propose an end-to-end model that explicitly learns to align unseen table schemas with seen ones. An overview of the architecture is shown in figure \ref{fig:alignnet}. As illustrated, the main difference between our model and the traditional end-to-end table-to-sequence model lies on the encoder side.
            
            First, for each input table, the model randomly samples a support set $S = \{T^S_1, T^S_2, \cdots, T^S_k \}$ from the train set and encodes attribute representations $\mathbf{v}_{i_j^{S}}$ for each table $T^S_i$ in $S$. Then, the model computes an alignment score $r$ for each table using  algorithm \ref{alg:align}. The core step of algorithm \ref{alg:align} is the application of the Hungarian Algorithm \cite{doi:10.1002/nav.3800020109} designed for bipartite graph matching. It is based on the property that if a number is added or subtracted from all of the elements of one row or one column, the optimal alignment of the resulting matrix is still an optimal alignment of the original matrix. By reducing some matrix elements to zeros and keeping other elements negative, the algorithm searches for an optimal alignment within the positions of zeros. The searching process is performed by iterating over rows and columns. Note that the cosine similarity between two vectors can be negative and the last step allows mismatch when no satisfactory alignments could be found for an input attribute.

            After we find the highest alignment score $r$ and its associated best alignment $\{(i,j_i)\}_{i=1}^{n}$, attribute representation $\mathbf{e}^a_{i}$ of the input table is then replaced by attribute representation $\mathbf{e}_{j_i}^{S,a}$ from the support set schema. The later steps then follow the traditional table-to-sequence framework discussed in the previous section.
            
        \subsection{Learning}
        
            Our model is trained in an end-to-end fashion to maximize the log-likelihood of the gold output text sequence. The negative log-likelihood loss for a single training instance $(T,\mathbf{y}) \in \mathcal{D}$ is defined as:
            
            \begin{equation}
                L_{\mathrm{NLL}} = -\log p(\mathbf{y}|T, \boldsymbol{\theta})
            \end{equation}
            where $\boldsymbol{\theta}$ denotes the model parameters.
            
            Meanwhile, in order to provide guidance for better schema alignment, an additional loss term that take an alignment score $r$ into consideration is adopted. The final loss is denoted as 
            \begin{equation}
                L = \sum_{\mathcal{D}}  \left( L_{\mathrm{NLL}} + \lambda r L_{\mathrm{NLL}} \right).
            \end{equation}
            where $\lambda$ is a positive scalar hyperparameter.
            
            The rationale of the modified loss term is to minimize the gradient when an input table is aligned to an unsatisfactory schema with relatively low alignment score. On the other hand, when $L_{\mathrm{NLL}}$ is relatively high (which means the aligned schema cannot help to generate the expected output sequence), the loss term encourages the model to yield a lower alignment score.

\section{Experiments}

        \begin{table*}[!t]
            \centering
            \begin{tabular}{lccccccccccc}
            \thickhline
            \multicolumn{1}{c}{\multirow{2}{*}{Model}} & \multicolumn{2}{c}{50} &  & \multicolumn{2}{c}{100} &  & \multicolumn{2}{c}{200} &  & \multicolumn{2}{c}{500} \\ \cline{2-3} \cline{5-6} \cline{8-9} \cline{11-12} 
            \multicolumn{1}{c}{}                       & \textbf{Test}       & \textbf{Dev}       &  & \textbf{Test}       & \textbf{Dev}        &  & \textbf{Test}       & \textbf{Dev}        &  & \textbf{Test}       & \textbf{Dev}        \\ \hline
            Base                                       & 14.1      & 14.5     &  & 14.8      & 14.3      &  & 16.0      & 16.1      &  & 17.7      & 18.3      \\ \hline
            Base+targ-copy                             & 13.2      & 13.3     &  & 15.7      & 15.5      &  & 15.7      & 15.5      &  & 18.6      & 17.9     \\ \hline
            Base+MAML                                       & 9.9       & 9.8      &  & 11.7      & 12.0      &  & 16.6      & 16.3      &  & 18.4      & 18.7      \\ \hline
            AlignNet                            & \textbf{18.8}      & \textbf{19.3}     &  & \textbf{18.7}      & \textbf{18.8}      &  & \textbf{18.7}      & \textbf{18.5}      &  & \textbf{21.0}      & \textbf{20.6}      \\ \thickhline
            \end{tabular}
            \caption{Comparison of results on the benchmark dataset. We report BLEU-4 scores on training set sizes of 50, 100, 200, 500 with unseen attribute proportion above 80\%. The results are averaged over 10 randomly sampled train sets for each size.}\label{tab:unseenres}
        \end{table*}
        
        \begin{table*}[ht]
            \centering
            \begin{tabular}{lccccccccccc}
            \thickhline
            \multicolumn{1}{c}{\multirow{2}{*}{Model}} & \multicolumn{2}{c}{50} &  & \multicolumn{2}{c}{100} &  & \multicolumn{2}{c}{200} &  & \multicolumn{2}{c}{500} \\ \cline{2-3} \cline{5-6} \cline{8-9} \cline{11-12} 
            \multicolumn{1}{c}{}                       & \textbf{Test}       & \textbf{Dev}       &  & \textbf{Test}       & \textbf{Dev}        &  & \textbf{Test}       & \textbf{Dev}        &  & \textbf{Test}       & \textbf{Dev}        \\ \hline
            Base                                       & 18.9      & 19.0     &  & 23.2      & 23.3      &  & 22.3      & 23.8      &  & 23.7      & 25.5      \\ \hline
            Base+targ-copy                             & 12.9      & 13.0     &  & 12.5      & 12.0      &  & 16.0      & 16.2      &  & 22.0      & 21.7      \\ \hline
            Base+MAML                                       & 10.7      & 10.8     &  & 16.6     & 17.1      &  & 21.7      & 22.5      &  & \textbf{26.7}      & 25.5      \\ \hline
            AlignNet                            & \textbf{23.4}      & \textbf{22.1}     &  & \textbf{24.8}      & \textbf{25.0}      &  & \textbf{26.3}      & \textbf{25.4}      &  & 26.5     & \textbf{25.6}      \\ \thickhline
            \end{tabular}
            \caption{Comparison of results on the datasets with traditional settings that directly samples from the train set. In the same vein as table \ref{tab:unseenres}, we report the average scores over 10 randomly sampled for each training set size.}\label{tab:seenres}
        \end{table*}

    In this section, we first describe a benchmark dataset that we construct for the new task of table-to-text generation for unseen schemas. Then, we show the experimental results of our model and several other baselines on the benchmark dataset. Some qualitative analysis will be provided at last.

    \subsection{Dataset}
        Our benchmark dataset is constructed based on the \textsc{wikitabletext} dataset \cite{AAAI1816138}. \textsc{wikitabletext} is an open-domain table-to-text dataset collected from the whole Wikipedia, which means the table schemas are not restricted to any specific domains. Table \ref{tab:wikitabletextstats} shows some important statistics of this dataset. Compared with previous close-domain  datasets such as \textsc{robocup} \cite{Chen:2008:LST:1390156.1390173}, \textsc{rotowire} \cite{wiseman-etal-2017-challenges} and \textsc{wikibio} \cite{lebret-etal-2016-neural}, \textsc{wikitabletext} has the most diverse attribute types that are suitable to test the generalizability of table-to-text models.
        
        \begin{table}[ht]
            \centering
            \begin{tabular}{ll} \thickhline
                \textbf{Type}  & \textbf{Value} \\ \hline
                \#Instances  & 13.3K  \\ 
                \#Tokens & 185.0K \\  
                \#Attribute Types & 3.0K \\ 
                Average Length &  13.9 \\  
                \#Attribute per Table & 4.1 \\ \thickhline
            \end{tabular}
            \caption{Statistics of the \textsc{wikitabletext} dataset.}\label{tab:wikitabletextstats}
        \end{table}

        Using the original train/development/test split setup of \textsc{wikitabletext}, about 98\% of the attribute types of development set and test set are seen during training. Thus, we keep the original development and test set and subsample the train set in order to limit the number of seen attribute types. To be more specific, we sample train set of size 50, 100, 200, and 500 while keeping the proportion of unseen attribute types in development and test set to be over 80\%. To rule out the impact of randomness with regard to the choice of training instances, we randomly sampled 10 sets for each size.

    \subsection{Implementation Details}
         To deal with the problem of out-of-vocabulary words, we use 50-dimensional pretrained GloVe\footnote{We use the GloVe vector pretrained with 6B corpus https://nlp.stanford.edu/projects/glove/} word vector \cite{pennington-etal-2014-glove} as pretrained token embeddings. During training, we freeze the embeddings of words to maintain the semantics and allow other neural network parameters to be trainable. The hidden state size of $\mathrm{FFN}_e$ is set as 100. For the sequence-to-vector encoder, we use a bi-directional LSTM with a hidden state size of 100. We set the decoder hidden state size to be 200 and output token embedding size to be 25. The support set size is set to 25 for train set size 50 and 50 for train set size 100, 200, and 500. As for the vocabulary, we aim to generate words that appear more than 10 times in the training set. We use AdaDelta optimizer to adaptively change the learning rate from $1.0$. The hyperparameter in the loss term $\lambda$ is set to $0.01$.
        
        During decoding, we feed a special token $\langle s \rangle$ into the decoder in the beginning. We stop the generation process when a special ending token $\langle e \rangle$ is output or the length of the sequence exceeds 20. We apply beam search \cite{Sutskever:2014:SSL:2969033.2969173,bahdanau2014neural} of size 5 for all the models. In terms of the evaluation metric, we use BLEU-4 \cite{papineni-etal-2002-bleu} which is a widely adopted metric in natural language generation tasks. 
        
    \subsection{Baselines}
        \subsubsection{Base} 
        We first implemented the state-of-the-art model from \cite{AAAI1816138} and denote it as Base. The architecture of Base is the same with as the one shown in figure \ref{fig:alignnet} without schema aligner module.
        
        \subsubsection{Base+targ-copy}
        Inspired by the work of \citeauthor{Hashimoto:2018:RFP:3327546.3327670}, we implement a model that retrieves the most similar instance from train set and generates a textual sequence $\mathbf{y}$ conditioning on the input table $T$ and retrieved instance $(T^\star,\mathbf{y}^\star)$. The model is implemented with extended copy mechanism that could copy from $\mathbf{y}^\star$.
        
        \subsubsection{Base+MAML}
        Inspired by the success of applying meta-learning method to text-to-SQL task \cite{huang-etal-2018-natural}, we implemented a MAML-based \cite{Finn:2017:MMF:3305381.3305498} framework for table-to-text NLG with pseudo-tasking \cite{huang-etal-2018-natural}. We use bag-of-embedding of attributes to calculate similarity scores between tables and construct a pseudo-task that consists of top-5 similar instances for each input instance. The pseudo-tasks then serve as the support sets of MAML.
        
    \subsection{Experimental Results}

        We conduct experiments on the benchmark dataset introduced previously. Models are trained on train sets of different sizes and evaluated on the original development and test set. Following previous work, we use BLEU-4 as an automatic evaluation metric. Evaluation results averaged over 10 randomly sampled sets are reported in table \ref{tab:unseenres}. 
        
        In order to get a better understanding of the challenges presented by the proposed task, we furthermore compare it with a conventional setup which puts no limits on unseen attribute proportion. Training dataset sizes are set to 50, 100, 200, 500 without controlling the number of unseen attribute types. All the results listed in table \ref{tab:seenres} are averaged over 10 randomly sampled training datasets of each size. 
        
        First of all, it can be seen from the results that our AlignNet model gives better or comparable performance other baseline models on most of the datasets not only under the unseen schema settings but also the traditional settings. Second, the comparison between the results of table \ref{tab:unseenres} and table \ref{tab:seenres} shows that the prevalence of unseen schemas during testing brings more challenges than traditional table-to-text settings to the models. Moreover, it can be seen from the results that AlignNet outperforms other baselines by a large margin when the train set size is 50, 100, 200 and 500 (33.3\%, 26.3\%, 16.8\%, 18.6\%) relative improvement under unseen settings, 23.8\%, 6.8\%, 17.9\%, 11.8\% under traditional settings).

    \subsection{Ablation Study}
    
        \begin{figure*}[t]
            \centering
            \includegraphics[scale=0.16]{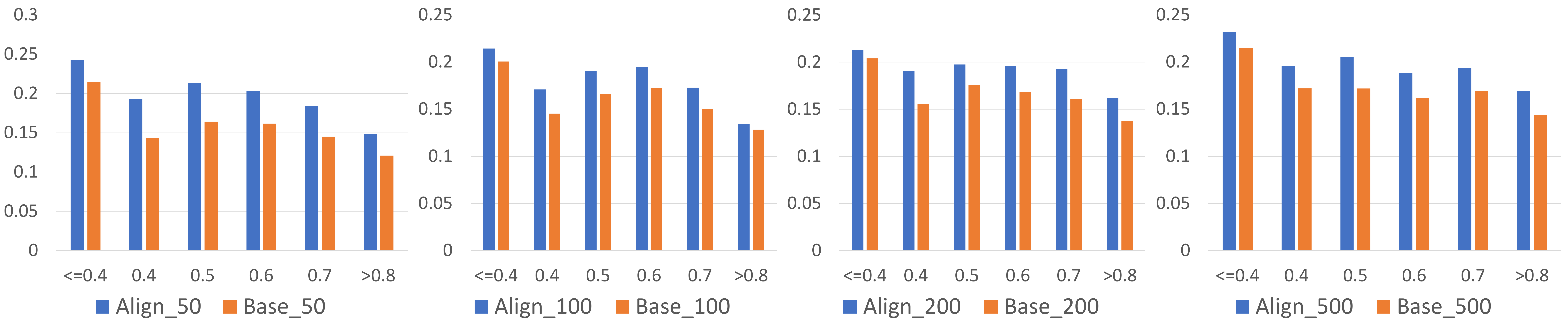}
            \caption{Development set BLEU-4 scores by varying different unseen attribute proportions. We calculate the average score of instances with certain unseen attribute portions. The X axis is unseen attribute proportion and the Y axis is the BLEU-4 score. For example, on the X axis, 0.5 refers to the average score of all development set instances with 50\%-60\% unseen attributes.}
            \label{fig:byportion}
        \end{figure*}
        \begin{figure*}[t]
            \centering
            \includegraphics[scale=0.16]{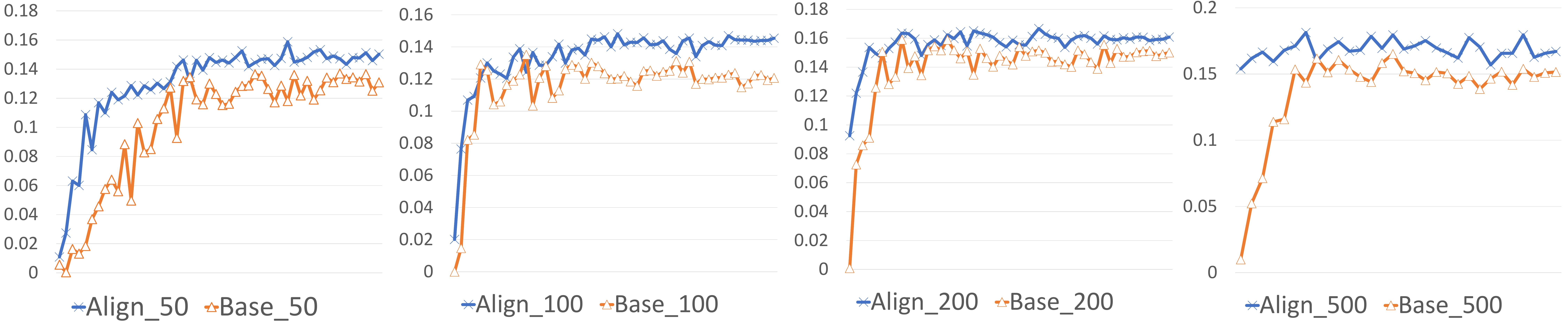}
            \caption{Learning curves (BLEU-4) of the first 50 epochs of the Base model and AlignNet.}
            \label{fig:learningcurve}
        \end{figure*}

        \subsubsection{Performance on Different Unseen Attribute Proportions}
            To examine how the amount of unseen attributes in a table affects the performance of our model, we report the performance by various unseen proportions. As demonstrated in figure \ref{fig:byportion}, the AlignNet model shows the most noticeable improvements compared with the Base model for input table schemas with 40\%-70\% unseen attributes. At the same time, limited improvement is shown for instances with less than 40\% unseen attributes, which indicates that the AlignNet works best for tables with a moderate proportion of unseen attribute types.

        \subsubsection{Learning Curve}
            Figure \ref{fig:learningcurve} shows the learning curves of the AlignNet and the Base model on the development set. We plot the averaged BLEU-4 scores over 10 sampled datasets of each size for the first 50 training epochs. As illustrated in figure \ref{fig:learningcurve}, AlignNet converges faster than the Base model. We hypothesize that the AlignNet better leverages the information of the training set by explicitly learning an alignment between schemas and directly copy the representations of attributes. Thus, AlignNet requires much smaller efforts to learn a generalizable representation for unseen schemas.

    \subsection{Qualitative Analysis}
        \subsubsection{Generation Example}
            \begin{table}[!t]
                \centering
                \begin{tabular}{ll}
                \thickhline
                \textbf{Attribute}                         & \textbf{Value}                                         \\ \hline
                $\star$Year                                       & 1985                                          \\
                $\star$Winner                                     & Peter Glover                                 \\
                $\star$Car                                   & Cheetah mk 8 Volkswagen                                       \\
                Team                               & Peter Macrow                                     \\ \hline
                \multicolumn{2}{l}{\makecell[l]{\textbf{Gold reference}: Peter Glover was from team \\ Peter Macrow.}} \\ \thickhline
                \multicolumn{2}{l}{\makecell[c]{Generated texts}} \\ \hline
                \multicolumn{2}{l}{\makecell[l]{\textbf{Base}: Peter Glover was \textless unk \textgreater  \textless unk \textgreater \\ Peter Macrow.}}            \\ \hline
                \multicolumn{2}{l}{\makecell[l]{\textbf{AlignNet}: Peter Glover was from team \\ Peter Macrow.}}   \\   \thickhline
                \end{tabular}
                
                \bigskip
                
                \begin{tabular}{ll}
                \thickhline
                \textbf{Attribute}              & \textbf{Value}                                         \\ \hline
                $\star$Name                                       & Ivan Cleary                                          \\
                $\star$Seasons                                     & 1996-1999                     \\
                $\star$Points                                   &  722                                      \\ \hline
                \multicolumn{2}{l}{\makecell[l]{\textbf{Gold reference}: Ivan Cleary got 722 points \\ during 1996-1999 season.}} \\ \thickhline
                \multicolumn{2}{l}{\makecell[c]{Generated texts}} \\ \hline
                \multicolumn{2}{l}{\makecell[l]{\textbf{Base}: 722 \textless{}unk\textgreater{} 722 in 1996-1999.}}            \\ \hline
                \multicolumn{2}{l}{\makecell[l]{\textbf{AlignNet}: In 1996-1999 season, Ivan Cleary \\ got 722 points.}}   \\   \thickhline
                \end{tabular}
                \caption{Comparison of generated samples between our AlignNet and the Base model. `$\star$' denotes unseen attribute types during training.}\label{tab:qualitative}
            \end{table}

            \begin{table}[!ht]
                \centering
                
                \begin{tabular}{ll}
                \thickhline
                \textbf{Test Attribute} & \textbf{Train Attribute} \\ \hline
                 $\star$Year  & Season    \\
                 $\star$Winner  & Rider \\
                $\star$Car & - \\
                 Team &  Team \\ \hline
                 $\star$Name  &  Player   \\
                 $\star$Seasons  & Season \\
                 $\star$Points & Goals \\ \hline
                \thickhline
                \end{tabular}
                \caption{Aligned train set schemas found by AlignNet for the two examples in \ref{tab:qualitative}. `$\star$' denotes unseen attribute types during training.}\label{tab:alignments}
            \end{table}

            Table \ref{tab:qualitative} shows two generated examples of the AlignNet and the Base model trained with a dataset of size 200. Due to the presence of unseen attributes, the Base model fails to generate correct textual descriptions for the input tables and tends to generate out-of-vocabulary tokens ($\textless{}unk\textgreater{}$). While the AlignNet correctly selects the contents that needs need to be said and verbalizes them in a sensible way.

        \subsubsection{Schema Alignment Example}
            
            Additionally, we show the schema alignments yielded for the generated examples by the AlignNet model in table \ref{tab:alignments}. For the first input, the model successfully finds a schema consists of attribute ``season'' and ``rider'' and matches them with unseen attributes ``year'' and ``winner''. Since we allow mismatch during aligning, attribute ``car'' has no corresponding attribute and its representation is, therefore, left unchanged. For the second input, all the unseen attributes are aligned to seen ones that represent similar content fields.

\section{Conclusion and Future Work}
    In this paper, we propose the novel task of table-to-text generation for unseen schemas which especially focuses on testing the ability to generalize. In order to solve the problem of unseen schemas, we propose the AlignNet which explicitly aligns unseen schemas to seen ones in the train set to get a better representation of the table. To evaluate the performance of different methods on this new task, we construct a benchmark dataset and conduct extensive experiments.
    
    In future work, we intend to explore more structural information such as latent categorical variables and co-occurrence of attribute types that lies in the table, which is a promising direction towards more generalizable table-to-text systems.

\bibliography{acl2020}

\begin{thebibliography}{40}
\expandafter\ifx\csname natexlab\endcsname\relax\def\natexlab#1{#1}\fi

\bibitem[{Angeli et~al.(2010)Angeli, Liang, and
  Klein}]{angeli-etal-2010-simple}
Gabor Angeli, Percy Liang, and Dan Klein. 2010.
\newblock \href {https://www.aclweb.org/anthology/D10-1049} {A simple
  domain-independent probabilistic approach to generation}.
\newblock In \emph{EMNLP}, pages 502--512, Cambridge, MA. Association for
  Computational Linguistics.

\bibitem[{Bahdanau et~al.(2014)Bahdanau, Cho, and Bengio}]{bahdanau2014neural}
Dzmitry Bahdanau, Kyunghyun Cho, and Yoshua Bengio. 2014.
\newblock Neural machine translation by jointly learning to align and
  translate.
\newblock \emph{arXiv preprint arXiv:1409.0473}.

\bibitem[{Bao et~al.(2018)Bao, Tang, Duan, Yan, Lv, Zhou, and
  Zhao}]{AAAI1816138}
Junwei Bao, Duyu Tang, Nan Duan, Zhao Yan, Yuanhua Lv, Ming Zhou, and Tiejun
  Zhao. 2018.
\newblock \href {https://aaai.org/ocs/index.php/AAAI/AAAI18/paper/view/16138}
  {Table-to-text: Describing table region with natural language}.

\bibitem[{Belz(2008)}]{belz_2008}
Anja Belz. 2008.
\newblock \href {https://doi.org/10.1017/S1351324907004664} {Automatic
  generation of weather forecast texts using comprehensive probabilistic
  generation-space models}.
\newblock \emph{Natural Language Engineering}, 14(4):431–455.

\bibitem[{Cahill and van Genabith(2006)}]{Cahill:2006:RPG:1220175.1220305}
Aoife Cahill and Josef van Genabith. 2006.
\newblock \href {https://doi.org/10.3115/1220175.1220305} {Robust pcfg-based
  generation using automatically acquired lfg approximations}.
\newblock In \emph{ACL}, ACL-44, pages 1033--1040, Stroudsburg, PA, USA.
  Association for Computational Linguistics.

\bibitem[{Chen and Mooney(2008)}]{Chen:2008:LST:1390156.1390173}
David~L. Chen and Raymond~J. Mooney. 2008.
\newblock \href {https://doi.org/10.1145/1390156.1390173} {Learning to
  sportscast: A test of grounded language acquisition}.
\newblock In \emph{Proceedings of the 25th International Conference on Machine
  Learning}, ICML '08, pages 128--135, New York, NY, USA. ACM.

\bibitem[{Dale et~al.(2003)Dale, Geldof, and
  Prost}]{Dale:2003:CUN:783106.783111}
Robert Dale, Sabine Geldof, and Jean-Philippe Prost. 2003.
\newblock \href {http://dl.acm.org/citation.cfm?id=783106.783111} {Coral: Using
  natural language generation for navigational assistance}.
\newblock In \emph{Proceedings of the 26th Australasian Computer Science
  Conference - Volume 16}, ACSC '03, pages 35--44, Darlinghurst, Australia,
  Australia. Australian Computer Society, Inc.

\bibitem[{Elsahar et~al.(2018)Elsahar, Gravier, and
  Laforest}]{elsahar-etal-2018-zero}
Hady Elsahar, Christophe Gravier, and Frederique Laforest. 2018.
\newblock \href {https://doi.org/10.18653/v1/N18-1020} {Zero-shot question
  generation from knowledge graphs for unseen predicates and entity types}.
\newblock In \emph{NAACL HLT}, pages 218--228, New Orleans, Louisiana.
  Association for Computational Linguistics.

\bibitem[{Finn et~al.(2017)Finn, Abbeel, and
  Levine}]{Finn:2017:MMF:3305381.3305498}
Chelsea Finn, Pieter Abbeel, and Sergey Levine. 2017.
\newblock \href {http://dl.acm.org/citation.cfm?id=3305381.3305498}
  {Model-agnostic meta-learning for fast adaptation of deep networks}.
\newblock In \emph{Proceedings of the 34th International Conference on Machine
  Learning - Volume 70}, ICML'17, pages 1126--1135. JMLR.org.

\bibitem[{Gardent et~al.(2017)Gardent, Shimorina, Narayan, and
  Perez-Beltrachini}]{gardent-etal-2017-webnlg}
Claire Gardent, Anastasia Shimorina, Shashi Narayan, and Laura
  Perez-Beltrachini. 2017.
\newblock \href {https://doi.org/10.18653/v1/W17-3518} {The {W}eb{NLG}
  challenge: Generating text from {RDF} data}.
\newblock In \emph{Proceedings of the 10th International Conference on Natural
  Language Generation}, pages 124--133, Santiago de Compostela, Spain.
  Association for Computational Linguistics.

\bibitem[{Gu et~al.(2016)Gu, Lu, Li, and Li}]{gu-etal-2016-incorporating}
Jiatao Gu, Zhengdong Lu, Hang Li, and Victor~O.K. Li. 2016.
\newblock \href {https://doi.org/10.18653/v1/P16-1154} {Incorporating copying
  mechanism in sequence-to-sequence learning}.
\newblock In \emph{ACL}, pages 1631--1640, Berlin, Germany. Association for
  Computational Linguistics.

\bibitem[{Hashimoto et~al.(2018)Hashimoto, Guu, Oren, and
  Liang}]{Hashimoto:2018:RFP:3327546.3327670}
Tatsunori~B. Hashimoto, Kelvin Guu, Yonatan Oren, and Percy Liang. 2018.
\newblock \href {http://dl.acm.org/citation.cfm?id=3327546.3327670} {A
  retrieve-and-edit framework for predicting structured outputs}.
\newblock In \emph{Proceedings of the 32Nd International Conference on Neural
  Information Processing Systems}, NIPS'18, pages 10073--10083, USA. Curran
  Associates Inc.

\bibitem[{Hu et~al.(2019)Hu, Xia, Neubig, and Carbonell}]{hu-etal-2019-domain}
Junjie Hu, Mengzhou Xia, Graham Neubig, and Jaime Carbonell. 2019.
\newblock \href {https://www.aclweb.org/anthology/P19-1286} {Domain adaptation
  of neural machine translation by lexicon induction}.
\newblock In \emph{ACL}, pages 2989--3001, Florence, Italy. Association for
  Computational Linguistics.

\bibitem[{Huang et~al.(2018)Huang, Wang, Singh, Yih, and
  He}]{huang-etal-2018-natural}
Po-Sen Huang, Chenglong Wang, Rishabh Singh, Wen-tau Yih, and Xiaodong He.
  2018.
\newblock \href {https://doi.org/10.18653/v1/N18-2115} {Natural language to
  structured query generation via meta-learning}.
\newblock In \emph{NAACL HLT}, pages 732--738, New Orleans, Louisiana.
  Association for Computational Linguistics.

\bibitem[{Kalchbrenner and
  Blunsom(2013)}]{kalchbrenner-blunsom-2013-recurrent-continuous}
Nal Kalchbrenner and Phil Blunsom. 2013.
\newblock \href {https://www.aclweb.org/anthology/D13-1176} {Recurrent
  continuous translation models}.
\newblock In \emph{EMNLP}, pages 1700--1709, Seattle, Washington, USA.
  Association for Computational Linguistics.

\bibitem[{Konstas and Lapata(2013)}]{Konstas:2013:GMC:2591248.2591256}
Ioannis Konstas and Mirella Lapata. 2013.
\newblock \href {http://dl.acm.org/citation.cfm?id=2591248.2591256} {A global
  model for concept-to-text generation}.
\newblock \emph{J. Artif. Int. Res.}, 48(1):305--346.

\bibitem[{Kuhn(1955)}]{doi:10.1002/nav.3800020109}
H.~W. Kuhn. 1955.
\newblock \href {https://doi.org/10.1002/nav.3800020109} {The hungarian method
  for the assignment problem}.
\newblock \emph{Naval Research Logistics Quarterly}, 2(1‐2):83--97.

\bibitem[{Lebret et~al.(2016)Lebret, Grangier, and
  Auli}]{lebret-etal-2016-neural}
R{\'e}mi Lebret, David Grangier, and Michael Auli. 2016.
\newblock \href {https://doi.org/10.18653/v1/D16-1128} {Neural text generation
  from structured data with application to the biography domain}.
\newblock In \emph{EMNLP}, pages 1203--1213, Austin, Texas. Association for
  Computational Linguistics.

\bibitem[{Li et~al.(2016)Li, Monroe, Ritter, Jurafsky, Galley, and
  Gao}]{li-etal-2016-deep}
Jiwei Li, Will Monroe, Alan Ritter, Dan Jurafsky, Michel Galley, and Jianfeng
  Gao. 2016.
\newblock \href {https://doi.org/10.18653/v1/D16-1127} {Deep reinforcement
  learning for dialogue generation}.
\newblock In \emph{EMNLP}, pages 1192--1202, Austin, Texas. Association for
  Computational Linguistics.

\bibitem[{Liu et~al.(2018)Liu, Wang, Sha, Chang, and Sui}]{AAAI1816599}
Tianyu Liu, Kexiang Wang, Lei Sha, Baobao Chang, and Zhifang Sui. 2018.
\newblock \href {https://aaai.org/ocs/index.php/AAAI/AAAI18/paper/view/16599}
  {Table-to-text generation by structure-aware seq2seq learning}.

\bibitem[{Ma et~al.(2019)Ma, Yang, Liu, Li, Zhou, and Sun}]{ma-etal-2019-key}
Shuming Ma, Pengcheng Yang, Tianyu Liu, Peng Li, Jie Zhou, and Xu~Sun. 2019.
\newblock \href {https://www.aclweb.org/anthology/P19-1197} {Key fact as pivot:
  A two-stage model for low resource table-to-text generation}.
\newblock In \emph{ACL}, pages 2047--2057, Florence, Italy. Association for
  Computational Linguistics.

\bibitem[{Moryossef et~al.(2019)Moryossef, Goldberg, and
  Dagan}]{moryossef-etal-2019-step}
Amit Moryossef, Yoav Goldberg, and Ido Dagan. 2019.
\newblock \href {https://doi.org/10.18653/v1/N19-1236} {{S}tep-by-step:
  {S}eparating planning from realization in neural data-to-text generation}.
\newblock In \emph{NAACL HLT}, pages 2267--2277, Minneapolis, Minnesota.
  Association for Computational Linguistics.

\bibitem[{Nie et~al.(2018)Nie, Wang, Yao, Pan, and
  Lin}]{nie-etal-2018-operation}
Feng Nie, Jinpeng Wang, Jin-Ge Yao, Rong Pan, and Chin-Yew Lin. 2018.
\newblock \href {https://doi.org/10.18653/v1/D18-1422} {Operation-guided neural
  networks for high fidelity data-to-text generation}.
\newblock In \emph{EMNLP}, pages 3879--3889, Brussels, Belgium. Association for
  Computational Linguistics.

\bibitem[{Novikova et~al.(2017)Novikova, Du{\v{s}}ek, and
  Rieser}]{novikova-etal-2017-e2e}
Jekaterina Novikova, Ond{\v{r}}ej Du{\v{s}}ek, and Verena Rieser. 2017.
\newblock \href {https://doi.org/10.18653/v1/W17-5525} {The {E}2{E} dataset:
  New challenges for end-to-end generation}.
\newblock In \emph{Proceedings of the 18th Annual {SIG}dial Meeting on
  Discourse and Dialogue}, pages 201--206, Saarbr{\"u}cken, Germany.
  Association for Computational Linguistics.

\bibitem[{Papineni et~al.(2002)Papineni, Roukos, Ward, and
  Zhu}]{papineni-etal-2002-bleu}
Kishore Papineni, Salim Roukos, Todd Ward, and Wei-Jing Zhu. 2002.
\newblock \href {https://doi.org/10.3115/1073083.1073135} {{B}leu: a method for
  automatic evaluation of machine translation}.
\newblock In \emph{ACL}, pages 311--318, Philadelphia, Pennsylvania, USA.
  Association for Computational Linguistics.

\bibitem[{Pennington et~al.(2014)Pennington, Socher, and
  Manning}]{pennington-etal-2014-glove}
Jeffrey Pennington, Richard Socher, and Christopher Manning. 2014.
\newblock \href {https://doi.org/10.3115/v1/D14-1162} {{G}love: Global vectors
  for word representation}.
\newblock In \emph{EMNLP}, pages 1532--1543, Doha, Qatar. Association for
  Computational Linguistics.

\bibitem[{Puduppully et~al.(2018)Puduppully, Dong, and
  Lapata}]{DBLP:journals/corr/abs-1809-00582}
Ratish Puduppully, Li~Dong, and Mirella Lapata. 2018.
\newblock \href {http://arxiv.org/abs/1809.00582} {Data-to-text generation with
  content selection and planning}.
\newblock \emph{CoRR}, abs/1809.00582.

\bibitem[{Qian and Yu(2019)}]{qian-yu-2019-domain}
Kun Qian and Zhou Yu. 2019.
\newblock \href {https://www.aclweb.org/anthology/P19-1253} {Domain adaptive
  dialog generation via meta learning}.
\newblock In \emph{ACL}, pages 2639--2649, Florence, Italy. Association for
  Computational Linguistics.

\bibitem[{Reiter and Dale(1997)}]{reiter_dale_1997}
Ehud Reiter and Robert Dale. 1997.
\newblock \href {https://doi.org/10.1017/S1351324997001502} {Building applied
  natural language generation systems}.
\newblock \emph{Natural Language Engineering}, 3(1):57–87.

\bibitem[{Reiter et~al.(2005)Reiter, Sripada, Hunter, Yu, and
  Davy}]{Reiter:2005:CWC:1113166.1644548}
Ehud Reiter, Somayajulu Sripada, Jim Hunter, Jin Yu, and Ian Davy. 2005.
\newblock \href {https://doi.org/10.1016/j.artint.2005.06.006} {Choosing words
  in computer-generated weather forecasts}.
\newblock \emph{Artif. Intell.}, 167(1-2):137--169.

\bibitem[{Rush et~al.(2015)Rush, Chopra, and Weston}]{rush-etal-2015-neural}
Alexander~M. Rush, Sumit Chopra, and Jason Weston. 2015.
\newblock \href {https://doi.org/10.18653/v1/D15-1044} {A neural attention
  model for abstractive sentence summarization}.
\newblock In \emph{EMNLP}, pages 379--389, Lisbon, Portugal. Association for
  Computational Linguistics.

\bibitem[{See et~al.(2017)See, Liu, and Manning}]{see-etal-2017-get}
Abigail See, Peter~J. Liu, and Christopher~D. Manning. 2017.
\newblock \href {https://doi.org/10.18653/v1/P17-1099} {Get to the point:
  Summarization with pointer-generator networks}.
\newblock In \emph{ACL}, pages 1073--1083, Vancouver, Canada. Association for
  Computational Linguistics.

\bibitem[{Shang et~al.(2015)Shang, Lu, and Li}]{shang-etal-2015-neural}
Lifeng Shang, Zhengdong Lu, and Hang Li. 2015.
\newblock \href {https://doi.org/10.3115/v1/P15-1152} {Neural responding
  machine for short-text conversation}.
\newblock In \emph{ACL-IJCNLP}, pages 1577--1586, Beijing, China. Association
  for Computational Linguistics.

\bibitem[{Sutskever et~al.(2014)Sutskever, Vinyals, and
  Le}]{Sutskever:2014:SSL:2969033.2969173}
Ilya Sutskever, Oriol Vinyals, and Quoc~V. Le. 2014.
\newblock \href {http://dl.acm.org/citation.cfm?id=2969033.2969173} {Sequence
  to sequence learning with neural networks}.
\newblock In \emph{Proceedings of the 27th International Conference on Neural
  Information Processing Systems - Volume 2}, NIPS'14, pages 3104--3112,
  Cambridge, MA, USA. MIT Press.

\bibitem[{Vinyals and Le(2015)}]{DBLP:journals/corr/VinyalsL15}
Oriol Vinyals and Quoc~V. Le. 2015.
\newblock \href {http://arxiv.org/abs/1506.05869} {A neural conversational
  model}.
\newblock \emph{CoRR}, abs/1506.05869.

\bibitem[{Wang et~al.(2018)Wang, Wu, Zhang, Su, and
  Wang}]{DBLP:journals/corr/abs-1811-02765}
Xin Wang, Jiawei Wu, Da~Zhang, Yu~Su, and William~Yang Wang. 2018.
\newblock \href {http://arxiv.org/abs/1811.02765} {Learning to compose
  topic-aware mixture of experts for zero-shot video captioning}.
\newblock \emph{CoRR}, abs/1811.02765.

\bibitem[{Wen et~al.(2016)Wen, Ga{\v{s}}i{\'c}, Mrk{\v{s}}i{\'c},
  Rojas-Barahona, Su, Vandyke, and Young}]{wen-etal-2016-multi}
Tsung-Hsien Wen, Milica Ga{\v{s}}i{\'c}, Nikola Mrk{\v{s}}i{\'c}, Lina~M.
  Rojas-Barahona, Pei-Hao Su, David Vandyke, and Steve Young. 2016.
\newblock \href {https://doi.org/10.18653/v1/N16-1015} {Multi-domain neural
  network language generation for spoken dialogue systems}.
\newblock In \emph{NAACL HLT}, pages 120--129, San Diego, California.
  Association for Computational Linguistics.

\bibitem[{Wiseman et~al.(2017)Wiseman, Shieber, and
  Rush}]{wiseman-etal-2017-challenges}
Sam Wiseman, Stuart Shieber, and Alexander Rush. 2017.
\newblock \href {https://doi.org/10.18653/v1/D17-1239} {Challenges in
  data-to-document generation}.
\newblock In \emph{EMNLP}, pages 2253--2263, Copenhagen, Denmark. Association
  for Computational Linguistics.

\bibitem[{Wong and Mooney(2007)}]{wong-mooney-2007-generation}
Yuk~Wah Wong and Raymond Mooney. 2007.
\newblock \href {https://www.aclweb.org/anthology/N07-1022} {Generation by
  inverting a semantic parser that uses statistical machine translation}.
\newblock In \emph{NAACL HLT}, pages 172--179, Rochester, New York. Association
  for Computational Linguistics.

\bibitem[{Zhao and Eskenazi(2018)}]{zhao-eskenazi-2018-zero}
Tiancheng Zhao and Maxine Eskenazi. 2018.
\newblock \href {https://doi.org/10.18653/v1/W18-5001} {Zero-shot dialog
  generation with cross-domain latent actions}.
\newblock In \emph{Proceedings of the 19th Annual {SIG}dial Meeting on
  Discourse and Dialogue}, pages 1--10, Melbourne, Australia. Association for
  Computational Linguistics.

\end{thebibliography}
\bibliographystyle{acl_natbib}

\end{document}